# Parallel Statistical and Machine Learning Methods for Estimation of Physical Load

Sergii Stirenko[1], Peng Gang[2], Wei Zeng[2], Yuri Gordienko[1*], Oleg Alienin[1], Oleksandr Rokovyi[1], Nikita Gordienko[1], Ivan Pavliuchenko[1], and Anis Rojbi[3]

[1]National Technical University of Ukraine "Igor Sikorsky Kyiv Polytechnic Institute", Kyiv, Ukraine
*yuri.gordienko@gmail.com
[2]Huizhou University, [2]Huizhou City, China
[3]CHArt Laboratory (Human and Artificial Cognitions), University of Paris 8, Paris, France

**Abstract.** Several statistical and machine learning methods are proposed to estimate the type and intensity of physical load and accumulated fatigue . They are based on the statistical analysis of accumulated and moving window data subsets with construction of a kurtosis-skewness diagram. This approach was applied to the data gathered by the wearable heart monitor for various types and levels of physical activities, and for people with various physical conditions. The different levels of physical activities, loads, and fitness can be distinguished from the kurtosis-skewness diagram, and their evolution can be monitored. Several metrics for estimation of the instant effect and accumulated effect (physical fatigue) of physical loads were proposed. The data and results presented allow to extend application of these methods for modeling and characterization of complex human activity patterns, for example, to estimate the actual and accumulated physical load and fatigue, model the potential dangerous development, and give cautions and advice in real time.

## 1 Introduction

Recently due to development of wearable electronics and Internet of Things, complex physiological signals can be registered including cerebral (electroencephalography, functional magnetic resonance imaging, etc.) and peripheral (heart rate, biological activity, temperature, etc.) ones [1-2]. They can be recorded and processed during various multimodal human-machine interactions. Quantitative characterization and interpretation of the mentioned physiological signals is non-trivial task which attracts attention of experts from various fields of science including medicine, biology, chemistry, electrical engineering, computer science, etc [3,4]. The main aim of this paper is to present the new approach to monitor and predict the type and level of current physical load by heart rate/beat analysis only (without accelerometry used in all other works) . The section 2.Background and Related Work gives the brief outline of the state of the art. The section 3.Experimental contains the description of the experimental part related with main terms, parameters and metrics. The section 4.Results reports about the results obtained and processed by some statistical and machine learning methods. The section 5.Discussion is dedicated to discussion of the results obtained and section 6.Conclusions summarizes the lessons learned.

## 2      Background and Related Work

Estimation of the actual physical load and fatigue is of great importance nowadays in the context of human-machine interactions, especially for health care and elderly care applications [3-6]. Evolution of information and communication technologies allows everyone to apply the range of the wearable sensors and actuators, which are already become de facto standard devices in the ordinary gadgets [1,7-9]. Recently several approaches of fatigue estimation were proposed on the basis of multimodal human-machine interaction and machine learning methods [4,10-14]. The valuable output can be obtained by usage of machine learning and, especially deep learning techniques, which are recently used for analysis of human physical activity [11-14]. During the last years various techniques were applied to measure physical load, stress, and fatigue by analysis of heart-rate, especially by measuring the RR interval or heart period variability [4,15-20]. Unfortunately, the type (walking, running, skiing, biking, etc.) and intensity (distance, time, pace, power, etc.) of the actual physical load hardly can be recognized by heart rate analysis only. Usually, to recognize them the heart monitors are used along with other wearable sensors like accelerometers, power meters, etc [21,22]. Moreover, any estimation of 'stress' and 'fatigue' should take into account the complex physiochemical and psychological state of humans under investigation, especially by heart rate analysis only. The more complicated wearable devices like EEG-monitors and brain-computer interfaces are applied for this purpose [23,24]. The data obtained often can be explained by various complex models including numerous parameters. Here the progress of the work is reported as to the new statistical method for characterization of the actual physical load by heart rate/beat analysis only with incentives for creation of some models describing the observed behaviors.

## 3      Experimental

The proposed statistical method is based on monitoring the human heart behavior, which can be estimated by heart beat/heart rate (HB/HR) monitors in the modern smartphones, smartwatches, fitness-trackers, or other fitness-related gadgets (like FitBit heart rate monitor, Armour39 heart rate monitor by Under Armour, etc.).

The typical example of the recorded HB/HR values is shown in Fig.1, where the previous rest phase is shown before the green vertical line with letter S for 'Start', the exercise itself (walking upstairs) is between the green and red vertical lines, and the following rest phase is located after the red vertical line with letter E for 'End'. The evident tendency is to have the high (very fluctuating part of the time series) HB/HR variability in the both rest states, and the much lower (the smoother part of the time series) HB/HR variability during exercise itself. This phenomenon is actively investigated to track and estimate stress states [4,15-20]. The main idea of the approach proposed here is to consider the time series of HB/HR values as statistical ensembles of values: a) accumulated from the beginning of the physical activity; b) contained inside a sliding timeslot window (for example, one hundred neighboring HB/HR measurements obtained by a sliding window). These ensembles are processed by calculation of mean, standard deviation, skewness, and kurtosis. Finally, these statistical parameters are plotted on the Pearson (kurtosis-skewness) diagram (see below in Fig.2), where kurtosis values are plotted versus square of skewness [25-28]. The similar approach in different ways was widely used for analysis of distributions and was successfully applied in various fields of science, including computer science, physics, materials science, finance, geoscience, etc. [29-33].

The following important aspects of this approach should be emphasized. The absolute HB/HR values are volatile and sensitive to its instant state (mood, stress, tremor, etc.) including momentary external disturbances and sources of noise. But the distributions of HB/HR values (cumulative or inside sliding timeslot window) and the statistical parameters of these distributions (mean, std, skewness, kurtosis) are not so volatile and can be more characteristic for the person itself (age, gender, physical maturity, fitness, accumulated fatigue, etc.) than for its instant state. In addition to this, the heart rate values are actually the integer values with 2-3 significant digits and not adequately characterize a heart activity (because the heart rate is actually the reverse value of the heartbeat multiplied by 60 seconds and rounded to integer value). In contrary, heartbeats are measured in milliseconds, contain 3-4 significant digits, and their usage gives at least10 times higher precision and more information.

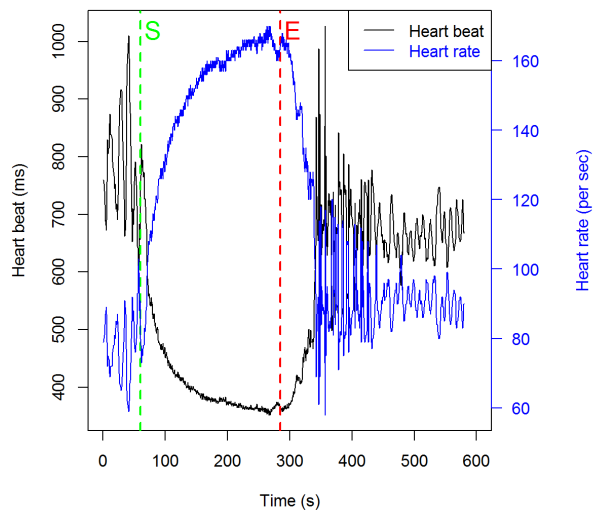

**Fig.**1.Time series of heart beat and heart rate values vs. exercise time for walking upstairs for the well-trained person (male, 47 years). The exercise was like: 1 min of rest (before the green vertical line with letter S for 'Start') + 3.45 min of walking upstairs (between the green and red vertical lines) + 5 min of rest (after the red vertical line with letter E for 'End').

The measurements of heart activity during exercises were performed by Armour39 heart rate monitor by Under Armour with the attachment point at breast. For the initial tests (feasibility study only) four male and female persons of various fitness (from beginners to marathoners) with age from 18 to 47 (mean weight 65±4 kg, mean height 1.71±0.05 m) were included in the study. All of them were volunteers and had not any known cardiac abnormalities. The next stages of this research will include the wider range of volunteers and these results will be reported separately elsewhere. The raw data were obtained for various physical activities in two experiments. The first experiment included analysis of HB distributions for walking upstairs, squats, dumbbells, push-ups (Section 4.1). The second experiment included analysis of influence of HR data on the models predicting the types of physical activities (Section 4.2).

## 4      Results

### 4.1    Heart Beat Distributions

The time series of HB values were obtained during various physical exercises and then they were considered as statistical samplings. Then the distributions of HB values in these samplings were analyzed (by calculation of mean, standard deviation, skewness, and kurtosis values) and plotted on the Pearson diagram (Fig.2).

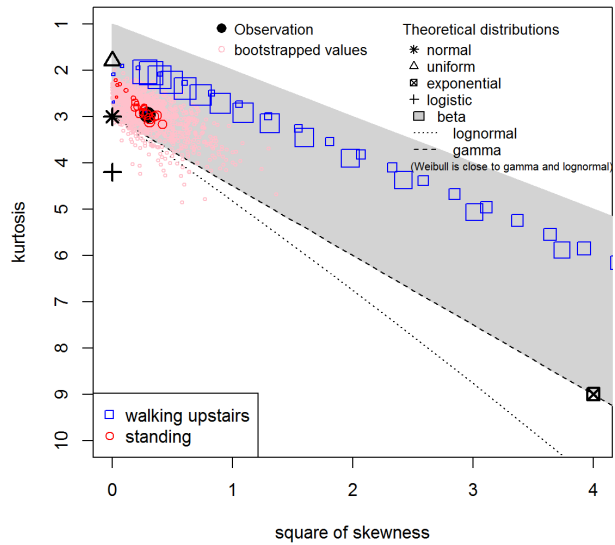

**Fig. 2.** The Pearson diagram for HB distributions vs. exercise time (the larger symbols correspond to the later times) for the well-trained person (male, 47 years). The exercise was like: 1 min of rest + 3.45 min of walking upstairs (13 floors) + 5 min of rest. Legend: The red circles denote the HB distributions in the initial standing position (near the normal distribution). The rose cloud of points denotes the results of bootstrapping analysis in the standing position. The blue rectangles denote the HB distributions during walking upstairs.

The example of raw data obtained for one of exercises (namely, for walking upstairs) is shown on the time series plot (Fig.1), and the example of the processed accumulated data is shown on the Pearson diagram (Fig.2). For the better visualization the size of symbol grows with the time of experiment and the bigger symbols corresponds to the later time moments. It allows us to observe the following tendency. Initially, in the preliminary rest state the distribution of HB values is close to the normal distribution. Then with the start of the physical exercise (walking upstairs) the distribution of HB values moves away from the location of normal distribution (black asterisk in Fig.2), but confines itself in the region of beta-distributions (gray zone in Fig.2).

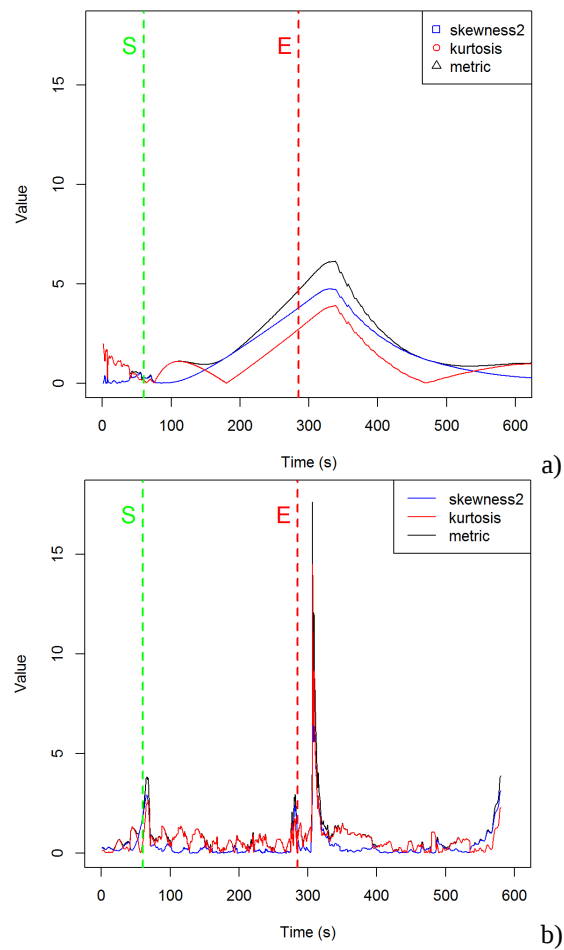

**Fig. 3.** Plots of statistical parameters (kurtosis and square of skewness) and Metric1 (the distance from the normal distribution on the Pearson diagram) of the HB distribution vs. exercise time: a) accumulated from the beginning of the physical activity; b) contained inside the sliding timeslot window.

After the end of exercises the HB distribution returns to the location of normal (black asterisk in Fig.2) and uniform (black triangle in Fig.2) distributions.

The dataset included numerous long time sequences (>$10^3$) and the statistical processing required numerous sub-samplings (~$10^3$) with many bootstrapping trials (>$10^3$ for each sub-sampling). That is why the several parallel processing techniques were implemented in R language designed for statistical analysis on various levels of granularity among different: a) time sequences, b) sub-samplings, and c) random subsets of sub-samplings for bootstrapping analysis.

The following metrics were proposed to characterize the accommodation and recovery levels during these exercises: on the Pearson diagram the distance from the normal distribution (Metric1) that corresponds to the rest state (where heart beats are not correlated) and the distance from the uniform distribution (Metric2) that corresponds to the increasing load on the heart (where heart beats grow with time). The plots in Fig.3 show evolution of the two metrics of the HB distributions in the experiment described above and shown in Fig.1 and Fig.2. For accumulated HB distribution (Fig.3a) the monotonous steady increase of Metric1 corresponds to the growing physical load, and the similar monotonous steady decrease does to the recovering during the rest after the end of the exercise (after the red vertical line with letter E) with some delay. For HB distribution in the sliding timeslot window (Fig.3b) the changes of the regime (start of exercise. end of exercise, recover) are followed by the sharp peaks of statistical parameters and the proposed Metric1. The slopes of Metric1 increase and decrease can be used also to characterize the accommodation and recovery levels during the exercises.

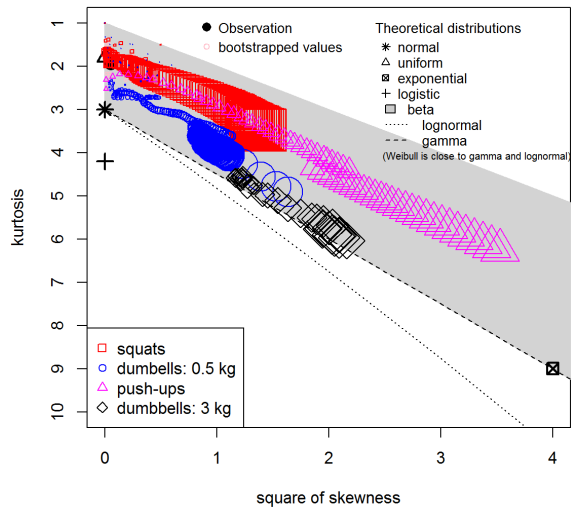

**Fig. 4.** The Pearson diagram for HB distributions vs. exercise time (the larger symbols correspond to the later times) for the well-trained person (male, 47 years). The exercises were like: squats, dumbbells of various weights (0.5 and 3 kg), push-ups. The active phases of the exercises without the rest and recovery stages are shown here.

The similar results were obtained for other types of exercises with various workloads: squats, dumbbells, push-ups (Fig.4). This method allows us to determine the level of workload even (compare location of HB distributions for dumbbells of various weight in Fig.4) and recovery rate after workloads.

### 4.2    Influence of HR Data on Models Predicting the Type of Physical Activities

At the moment the empirical results as to the change of distribution type for the accumulated HB values during physical exercise hardly have any simple explanations. But due to the recent success of various machine learning methods for analysis of the complex processes the incentive to apply some of them naturally appeared. The direct application of machine learning methods for the analysis of HB distributions is under work right now, but the preliminary similar tests on the simpler experiment are reported below.

The same focus group performed several physical activities of similar, but different types (actually running, skiing, and walking) with various intensities (distances and durations). The aim of experiment was to investigate feasibility to predict the type of physical activity from some dynamic features (distance, time, pace, velocity, etc.) and improve this prediction by adding heart activity features (average heart rate, maximal heart rate, etc.) by heart rate/beat analysis (not by accelerometry like in all other works) with some well-known methods including linear regression, neural network, and deep neural network. The features were divided in independent and derived. For example, distance and time are independent ones among the dynamic features, but pace (time/distance in minute/kilometer units), velocity (distance/time in meter/minute units), and MetricD (pace in square) are derived ones. Similarly, HR in rest (HRrest), maximal HR (MHR), minimal HR (minHR), average HR (AHR) are independent among the heart-related features, but working range HR (MHR-minHR), HR reserve (MHR - HRrest), and HR recovery (MHR - HRrest after exercise) are derived ones. To the moment the results are reported here on the following models:
model 1 (Fig. 5a) with *independent dynamic* features:

$$\textit{Type of Activity} \sim \textit{Distance} + \textit{Duration} \qquad (1)$$

model 2 (Fig. 5c) with *independent heart-related* features:

$$\textit{Type of Activity} \sim \textit{MHR} + \textit{AHR} \qquad (2)$$

model 3 (Fig. 5b) with *all* (independent and derived) *dynamic* features:

$$\textit{Type of Activity} \sim \textit{Distance} + \textit{Duration} + \textit{Pace} + \textit{Velocity} + \textit{MetricD} \qquad (3)$$

model 4 (Fig. 5d) with *all dynamic* features + *heart-related* ones:

$$\textit{Type of Activity} \sim \textit{Distance} + \textit{Duration} + \textit{Pace} + \textit{Velocity} + \textit{MetricD} + \textit{MHR} + \textit{AHR} \qquad (4)$$

The idea behind inclusion of derived features (like pace, velocity, MetricD) consists in inclusion of non-linear relations among independent parameters. On this stage of research, inclusion of heart-related features is limited to MHR and AHR, but usage of the more complex statistical parameters of HB/HR distributions (which are described in the previous subsection 4.1.Heart Beat Distributions) is under work now and will be reported elsewhere.

**Linear Regression.** The multiple linear regression model contains only one predictor variable (type of the physical activity), several explanatory variables (see above), and the relationship between the predictor variable and explanatory variables is assumed to be linear in this model. The results of application of the linear regression for various models are shown in Fig.5 for the standardized residuals (response minus fitted values) versus the fitted values (numbers near circles denote the numeration of exercises – only several of them are shown here). Some predicted types of the physical activities are demonstrated in Fig.8 and are discussed below in section 5.Discussion.

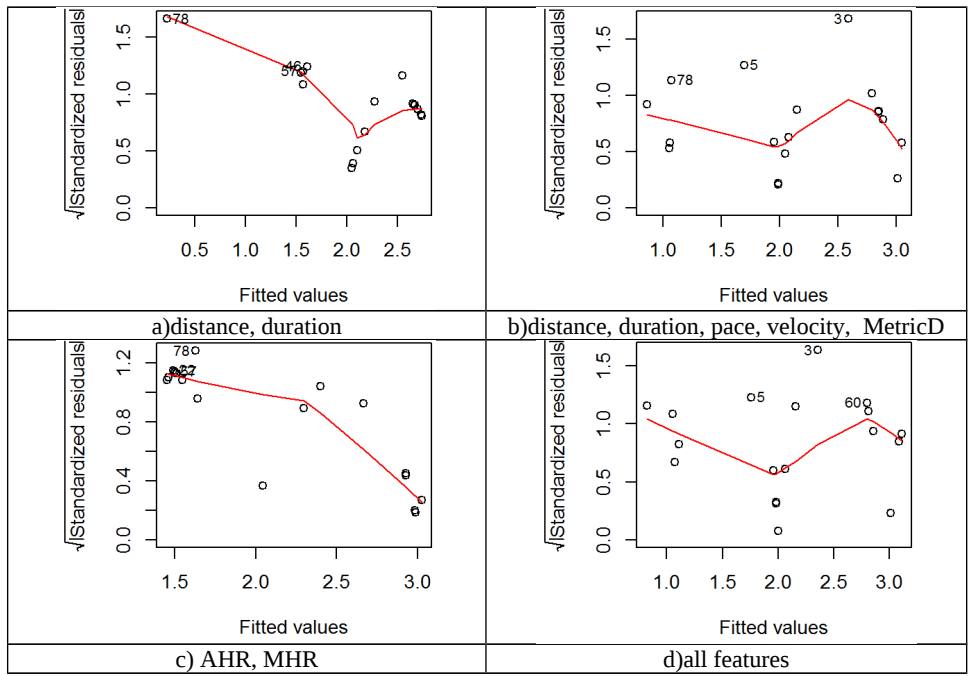

**Fig. 5.** Linear regression results: square root of standardized residuals vs. some fitted values.

**Neural Network.** The simple neural network (Fig.6) was used with various (depending on the model) input nodes (I), one output node (O) for the predicted value of the type of physical activity, 6 neurons (H1-H6) and biases (B). The dark lines mean positive values of weights, and light lines — negative ones. The widths of lines demonstrate the relative values of the weights, for example, contribution of the velocity is significant in models 1 (Fig. 5a) and 3 (Fig. 5b), that is graphically shown by the thicker lines. Again the examples of the predicted types of the physical activities are demonstrated in Fig.8 and are discussed below in section 5.Discussion.

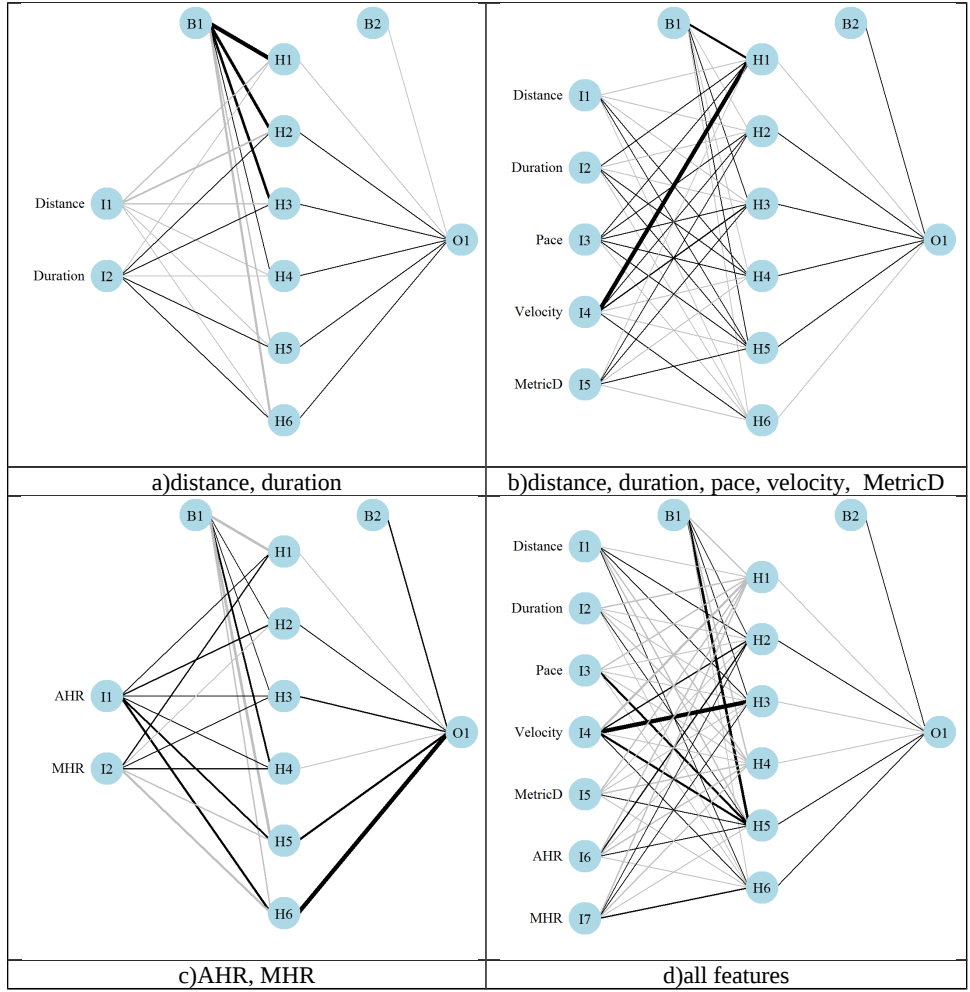

**Fig. 6.** Neural network with the contribution of some neurons graphically shown by the width of the lines (dark - positive weights, light - negative weights).

**Deep Neural Network.** The deep neural network (Fig.7) was used with various (depending on the model) input nodes (I), one output node (O) for the predicted value of the type of physical activity, 4 layers with (12, 8, 6, 3) neurons (H) and biases (B). Here the dark lines also mean positive values of weights, and light lines — negative ones after training the deep neural network with the tuned learning rate (0.001), the thresholds (0.001) for the partial derivatives of the error function as stopping criteria, logistic activation functions, and globally convergent algorithm based on the resilient backpropagation [34-36]. Here contribution of some features is not so evident like in the case of the shallow neural network from the previous subsection. The widths of lines demonstrate the relative values of the weights, and the contributions from neurons have the very complex patterns on all layers. For example in model 1 (with *independent dynamic* features) some intermediate hidden values propagated from the 1st layer to the 2nd layer contribute the most (Fig. 7a) while for model 4 (with *all dynamic* features + *heart-related* ones) most of the connections are almost equally important (Fig. 7d). At the moment these results are given for comparison with other machine learning methods (linear regression and neural network in previous subsections) and should be investigated thoroughly, especially in the view of hyper-parameter tuning that is planned to be performed, explained and discussed in the future work.

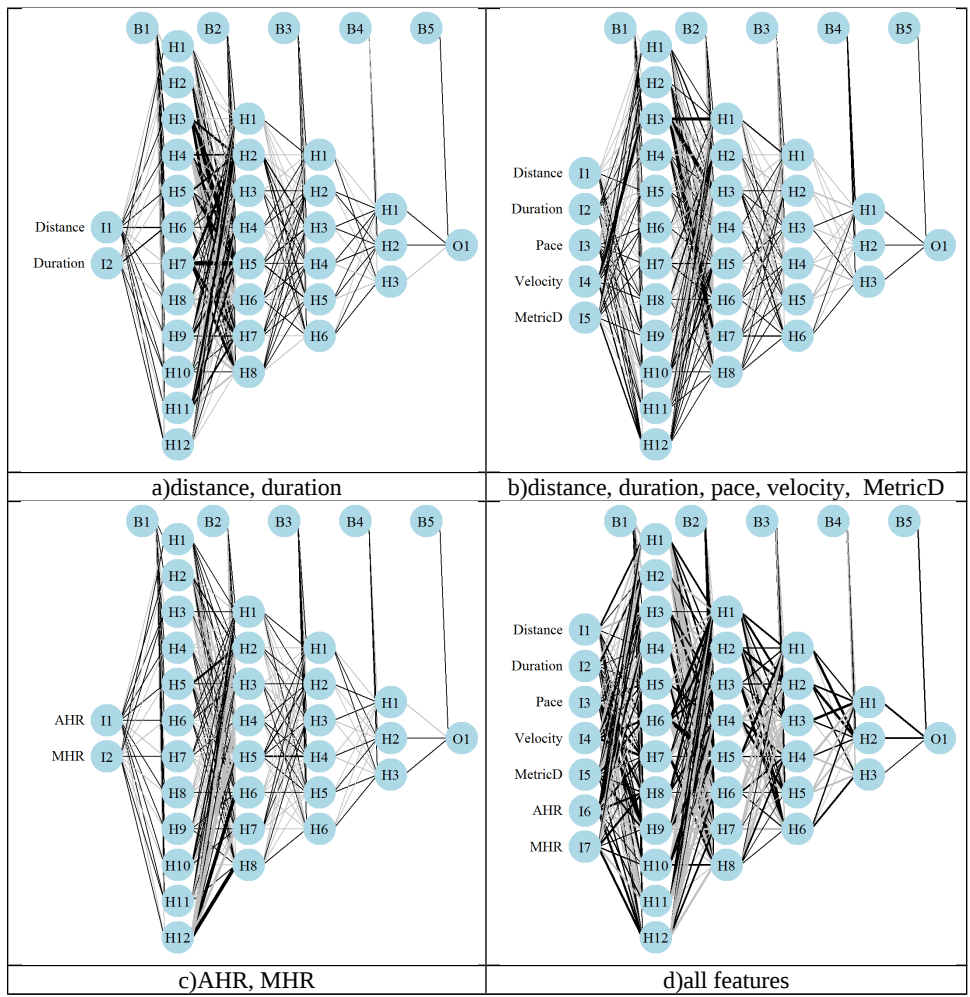

**Fig. 7.** Neural network with the contribution of some neurons graphically shown by the width of the lines (dark - positive weights, light - negative weights).

## 5   Discussion

Some types of the physical activities predicted by the above mentioned models are shown in Fig.8, where the codes for types of the physical activities are as follows: 1 — running, 2 — skiing, 3 — walking. The linear regression demonstrate the worst prediction abilities for all models, but addition of heart-related features in model 4 (Fig. 8d) slightly decrease the error and improve predictions for skiing and walking. In general, the larger models (with more input features) like model 3 (Fig. 8b) and model 4 (Fig. 8d) gives the better predictions than the smaller models like model 1 (Fig. 8a) and model 2 (Fig. 8c). But the most significant improvement of prediction was obtained for the shallow (1-layer) and deep (4-layer) neural networks in model 4 due to inclusion of the additional independent heart-related features.

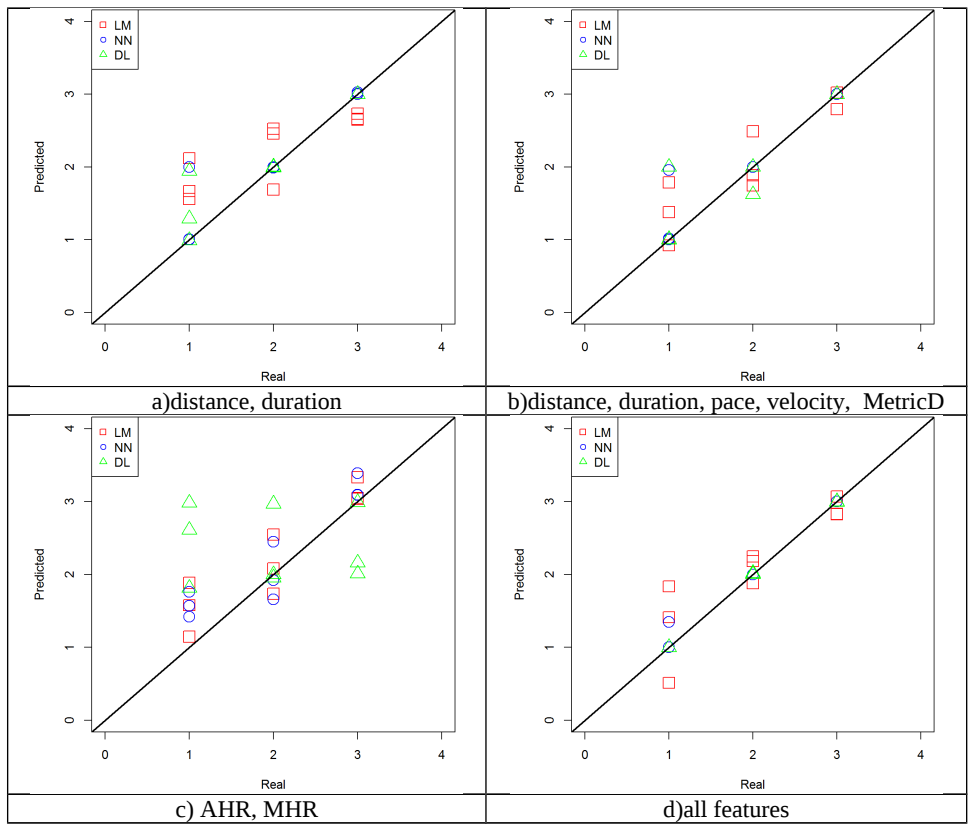

**Fig. 8.** Examples of the types of physical activities (1 — running, 2 — skiing, 3 — walking) predicted by linear regression (LM), shallow (NN) and deep learning (DL) neural network.

In fact, the increase of the number of independent input features should theoretically enlarge the information processed by neural networks (shallow and deep) and im-prove predictions. In this sense, inclusion of the heart-related features (even in such a limited way like addition of AHR and MHR) contribute the additional information about the physical exercise. But this contribution can be very sensitive to other aspects of the humans under tests, for example, typical for some focus groups (like age, gender, weight, physical maturity, etc.), and personal peculiarities (like fitness, mood, accumulated fatigue, etc.). On the one hand it does not allow to apply the models trained on some limited focus group for the wider range of users without understanding the influence of personal peculiarities. But on the other hand this influence of personal peculiarities can allow to create the personal models previously trained on the typical group and later tuned to the personal peculiarities of the concrete persons.

## 6     Conclusions

The type and intensity of current physical load is proposed to be monitored, estimated, and predicted by analysis of the time series of HB values considered as statistical ensembles of values: a) accumulated from the beginning of the physical activity; b) contained inside the sliding timeslot window. The mean, standard deviation, skewness, and kurtosis were calculated for these ensembles and plotted on the Pearson (kurtosis-skewness) diagram. The first experiment shown that location of statistical ensembles of HB values is close to the location of the normal distribution in rest, but it moves away with the time of exercises, and then return to the place of the normal distribution after the end of the exercises. This approach was applied to the data gathered by the wearable heart monitor for various types and levels of physical activities (walking upstairs, squats, dumbbells, push-ups) and the results demonstrated the same tendency. Several metrics for estimation of the instant effect and accumulated effect (physical fatigue) of physical loads were proposed. Inspired by the recent success of various machine learning models for analysis of the complex empirical data without the evident relationships, several machine learning models were proposed to analyze the observed complex behavior. The second experiment included the analysis of influence of HR data on the models predicting the types of physical activities. To the moment the limited set of features was used to predict the type of physical exercise (actually running, skiing, and walking) from dynamic features (distance, time, pace, velocity, etc.) and improve this prediction by adding heart activity features (average heart rate, maximal heart rate, etc.). Despite the limited dataset and short list of features the results of both experiments allow to conclude that statistical analysis of HB/HR time series as accumulated or sliding statistical distributions of HB/HR values can be a promising way for characterization of the actual physical load. Moreover the statistical parameters for distributions of HB/HR values (like maximum, minimum, normalized mean, standard deviation, skeweness, and kurtosis) can be used as additional independent features in shallow and deep neural networks.

The obtained predictions can be very sensitive to many additional collective (like age, gender, weight, physical maturity, etc.), and individual (like personal fitness, mood, accumulated fatigue, etc.) aspects of the humans under tests. That is why the much larger datasets and additional research will be necessary for the more collec-tively and personally tailored models. In this context, the further progress can be reached by sharing the similar datasets around the world in the spirit of open science, volunteer data collection, processing and computing [37-39]. The data and results presented allow us to extend application of these methods for modeling and characterization of complex human activity patterns. For example, under condition of the further improvement the models presented here can be used to estimate the actual and accumulated physical load and related fatigue, model the potential dangerous development, and give cautions and advice in real time, that is very important for many health and elderly care applications.

## Acknowledgment

The work was partially supported by Ukraine-France Collaboration Project (Programme PHC DNIPRO) (http://www.campusfrance.org/fr/dnipro) and by Huizhou Science and Technology Bureau and Huizhou University (Huizhou, P.R.China) in the framework of Platform Construction for China-Ukraine Hi-Tech Park Project # 2014C050012001.

## References

1. Kumari, P., Mathew, L., Syal, P.: Increasing trend of wearables and multimodal interface for human activity monitoring: A review. Biosensors and Bioelectronics 90, 298-307 (2017).
2. Koydemir, H. C., Ozcan, A.: Wearable and Implantable Sensors for Biomedical Applications. Annual Review of Analytical Chemistry 11, DOI: 10.1146/annurev-anchem-061417-125956 (2018).
3. Faust, O., Hagiwara, Y., Hong, T. J., Lih, O. S., & Acharya, U. R. Deep learning for healthcare applications based on physiological signals: a review. Computer Methods and Programs in Biomedicine (2018).
4. Mohanavelu, K., Lamshe, R., Poonguzhali, S., Adalarasu, K., & Jagannath, M. Assessment of Human Fatigue during Physical Performance using Physiological Signals: A Review. Biomedical and Pharmacology Journal, 10(4), 1887-1896 (2017).
5. Craighead, W. Edward, and Charles B. Nemeroff, eds. The concise Corsini encyclopedia of psychology and behavioral science, John Wiley & Sons (2004).
6. Gordienko, Y., Stirenko, S., Alienin, O., Skala, K., Sojat, Z., Rojbi, A., López Benito, J.R., Artetxe González, E., Lushchyk, U., Sajn, L., Llorente Coto, A., Jervan, G.: Augmented coaching ecosystem for non-obtrusive adaptive personalized elderly care on the basis of Cloud-Fog-Dew computing paradigm. In: 2017 40th International Convention on Information and Communication Technology, Electronics and Microelectronics (MIPRO), pp. 359-364. IEEE, Opatija, Croatia (2017).
7. Banaee, H., Ahmed, M. U., Loutfi, A. Data mining for wearable sensors in health monitoring systems: a review of recent trends and challenges. Sensors, 13(12), 17472-17500 (2013).
8. Bunn, J. A., Navalta, J. W., Fountaine, C. J., Reece, J. D.: Current State of Commercial Wearable Technology in Physical Activity Monitoring 2015–2017. International journal of exercise science, 11(7) 503 (2018).

9. Amft, O., & Van Laerhoven, K.: What Will We Wear After Smartphones?. IEEE Pervasive Computing, 16(4) 80-85 (2017).
10. Peng Gang, Jiang Hui , Stirenko, S., Gordienko, Y., Shemsedinov, T., Alienin, O., Kochura, Y., Gordienko, N., A.Rojbi, López Benito, J.R., González, E.A.: User-driven intelligent interface on the basis of multimodal augmented reality and brain-computer interaction for people with functional disabilities. Future of Information and Communications Conference (FICC), Singapore. arXiv preprint arXiv:1704.05915 (2017).
11. Du, L. H., Liu, W., Zheng, W. L., Lu, B. L.: Detecting driving fatigue with multimodal deep learning. In: 2017 8th International IEEE/EMBS Conference on Neural Engineering (NER) pp. 74-77. IEEE (2017).
12. Lopez, M. B., del-Blanco, C. R., Garcia, N.: Detecting exercise-induced fatigue using thermal imaging and deep learning. In: 2017 Seventh International Conference on Image Processing Theory, Tools and Applications (IPTA) pp. 1-6. IEEE (2017).
13. Gordienko, Y., Stirenko, S., Kochura, Y., Alienin, O., Novotarskiy, M., Gordienko, N., Rojbi, A.: Deep learning for fatigue estimation on the basis of multimodal human-machine interactions. XXIX IUPAP Conference in Computational Physics (CCP2017), Paris, France. arXiv preprint arXiv:1801.06048 (2017).
14. Hajinoroozi, M., Zhang, J. M., Huang, Y.: Driver's fatigue prediction by deep covariance learning from EEG. In: 2017 International Conference on Systems, Man, and Cybernetics (SMC) pp. 240-245. IEEE (2017).
15. Togo, F., & Takahashi, M.: Heart Rate Variability in Occupational Health—A Systematic Review. Industrial health 47(6), 589-602 (2009).
16. Aubert, A. E., Seps, B., Beckers, F.: Heart rate variability in athletes. Sports medicine, 33(12), 889-919 (2003).
17. Schmitt, L., Regnard, J., Desmarets, M., Mauny, F., Mourot, L., Fouillot, J. P., Coulmy, N., Millet, G.: Fatigue shifts and scatters heart rate variability in elite endurance athletes. PloS one 8(8), e71588 (2013).
18. Pichot, V., Roche, F., Gaspoz, J. M., Enjolras, F., Antoniadis, A., Minini, P., Costes, F., Busso, T., Lacour, J-R., Barthelemy, J. C.: Relation between heart rate variability and training load in middle-distance runners. Medicine and science in sports and exercise 32(10), 1729-1736 (2000).
19. Gonzalez, K., Sasangohar, F., Mehta, R. K., Lawley, M., Erraguntla, M.: Measuring Fatigue through Heart Rate Variability and Activity Recognition: A Scoping Literature Review of Machine Learning Techniques. In: Proceedings of the Human Factors and Ergonomics Society Annual Meeting 61(1), pp. 1748-1752. Sage CA: Los Angeles, CA: SAGE Publications (2017).
20. Morgan, S. J., Mora, J. A. M.: Effect of Heart Rate Variability Biofeedback on Sport Performance, a Systematic Review. Applied psychophysiology and biofeedback 42(3), 235-245 (2017).
21. Yang, C. C., Hsu, Y. L.: A review of accelerometry-based wearable motion detectors for physical activity monitoring. Sensors 10(8), 7772-7788 (2010).
22. Evenson, K. R., Goto, M. M., Furberg, R. D.: Systematic review of the validity and reliability of consumer-wearable activity trackers. International Journal of Behavioral Nutrition and Physical Activity 12(1), 159 (2015).
23. Lin, C. T., Ko, L. W., Chang, M. H., Duann, J. R., Chen, J. Y., Su, T. P., Jung, T. P.: Review of wireless and wearable electroencephalogram systems and brain-computer interfaces–a mini-review. Gerontology 56(1), 112-119 (2010).

24. Kumari, P., Mathew, L., Syal, P.: Increasing trend of wearables and multimodal interface for human activity monitoring: A review. Biosensors and Bioelectronics 90, 298-307 (2017).
25. Cramer, H., Mathematical Methods of Statistics, Vol. 9, Princeton University Press, Princeton (1999).
26. Delignette-Muller, M.L., Pouillot, R., Denis, J.-B., Dutang, C.: fitdistrplus package for R (2012).
27. Cullen, A., Frey, H.: Probabilistic Techniques in Exposure Assessment: A Handbook for Dealing with Variability and Uncertainty in Models and Inputs, Springer (1999).
28. Gordienko, Y. G.: Generalized model of migration-driven aggregate growth—asymptotic distributions, power laws and apparent fractality. International Journal of Modern Physics B 26(01), 1250010 (2012).
29. Ma, X., Xu, F.: Peak factor estimation of non‐Gaussian wind pressure on high‐rise buildings. The Structural Design of Tall and Special Buildings 26(17), e1386 (2017).
30. Gordienko, Y. G.: Molecular dynamics simulation of defect substructure evolution and mechanisms of plastic deformation in aluminium nanocrystals. Metallofizika i Noveishie Tekhnologii 33(9), 1217-1247 (2011).
31. Ketchantang, W., Derrode, S., Martin, L., Bourennane, S.: Pearson-based mixture model for color object tracking. Machine Vision and Applications 19(5-6), 457-466 (2008).
32. Tison, C., Nicolas, J. M., Tupin, F., Maître, H.: A new statistical model for Markovian classification of urban areas in high-resolution SAR images. IEEE transactions on geoscience and remote sensing 42(10), 2046-2057 (2004).
33. Sornette, D., Zhou, W. X.: Predictability of large future changes in major financial indices. International Journal of Forecasting, 22(1), 153-168 (2006).
34. Anastasiadis, A. D., Magoulas, G. D., Vrahatis, M. N.: New globally convergent training scheme based on the resilient propagation algorithm. Neurocomputing 64, 253-270 (2005).
35. Intrator O., Intrator N.: Using Neural Nets for Interpretation of Nonlinear Models. In: Proceedings of the Statistical Computing Section, pp. 244-249. American Statistical Society (eds), San Francisco (1993).
36. Beck, M.W.: Visualizing neural networks, https://github.com/fawda123, , last accessed 2018/05/03.
37. Goldberger, A. L., Amaral, L. A., Glass, L., Hausdorff, J. M., Ivanov, P. C., Mark, R. G., Mietus, J.E., Moody, G.B., Peng, C-K., Stanley, H. E.: Physiobank, physiotoolkit, and physionet. Circulation, 101(23), e215-e220 (2000).
38. Gordienko, N., Lodygensky, O., Fedak, G., Gordienko, Yu.: Synergy of volunteer measurements and volunteer computing for effective data collecting, processing, simulating and analyzing on a worldwide scale, In: Proc. 38th Int. Convention on Inf. and Comm. Technology, Electronics and Microelectronics (MIPRO), pp. 193-198. IEEE, Opatija, Croatia (2015).
39. Chen, Y., Wang, Z. Y., Yuan, G., Huang, L.: An overview of online based platforms for sharing and analyzing electrophysiology data from big data perspective. WIREs Data Mining and Knowledge Discovery 7(4) e1206 (2017).